\documentclass[10pt,twocolumn,letterpaper]{article}

\usepackage[pagenumbers]{cvpr} 

\usepackage{graphicx}
\usepackage{amsmath}
\usepackage{amssymb}
\usepackage{booktabs}
\usepackage{tabularx}
\makeatletter
\@namedef{ver@everyshi.sty}{}
\makeatother
\usepackage{tikz}
\usepackage{pgfplots,pgfplotstable}
\usepackage{enumitem}
\usepackage{multicol}
\usepackage{multirow}
\usepackage{amssymb}
\usepackage{dblfloatfix}
\usepackage{colortbl}
\usepgfplotslibrary{fillbetween}
\usepgfplotslibrary{colorbrewer}
\pgfplotsset{compat=1.16}
\usepackage{textcomp}

\usepackage[accsupp]{axessibility}

\usepackage[pagebackref,breaklinks,colorlinks]{hyperref}

\usepackage[capitalize]{cleveref}
\crefname{section}{Sec.}{Secs.}
\Crefname{section}{Section}{Sections}
\Crefname{table}{Table}{Tables}
\crefname{table}{Tab.}{Tabs.}

\def\cvprPaperID{22} 
\def\confName{CVPR}
\def\confYear{2023}

\newcolumntype{Y}{>{\centering\arraybackslash}X}

\begin{document}

\pgfplotstableread[row sep=\\,col sep=&]{
    metric & model1 & model2 & model3 & model4 & model5 & model6 & model7 & model8 & model9\\
    AUROC & 100 & 100 & 100 & 95.49 & 95.49 & 95.49 & 84.68 & 84.68 & 84.68 \\
    AUPR-in & 100 & 100 & 100  & 97.74 & 96.65 & 97.69 & 80.36 & 92.70 & 83.23 \\
    AUPR-out & 100 & 100 & 100  & 91.16 & 93.69 & 90.63 & 88.19 & 63.94 & 85.74 \\
    FPR95 & 0     & 0   & 0    & 15.9  & 17.82 & 17.75 & 60.80 & 46.34 & 62.61 \\
    FNR95 & 0     & 0   & 0    & 30.51  & 26.11 & 26.03 & 46.46 & 80.14 & 53.7 \\
    }\scores

\pgfplotstableread[row sep=\\,col sep=&]{
    metric & model1 & model2 & model3 & model4 & model5 & model6 & model7 & model8 & model9\\
    AUTC & 0.0517 & 0.4272 & 0.2947 & 0.4324 & 0.3454 & 0.1882 & 0.3236 & 0.3194 & 0.4095 \\
    AUFPR & 0.00949366908032816 & 0.27096639987366516 & 0.20213067343145172 & 0.009294902885467738 & 0.039859673235023915 & 0.17382973500595464 & 0.4872734194916606 & 0.24048643295811778 & 0.31345100662143544 \\
    AUFNR & 0.09393890818571535 & 0.5833620098400444 & 0.387304440570849 & 0.8555512000512685 & 0.6509759023678509 & 0.2026014813624371 & 0.15989627619339097 & 0.39822523101434787 & 0.5055782308120128 \\
    }\ourscores

\title{Beyond AUROC \& co. for evaluating\\out-of-distribution detection performance}


\author{Galadrielle Humblot-Renaux$^1$\qquad Sergio Escalera$^{1,2,3}$\qquad Thomas B. Moeslund$^1$\\%
$^1$Visual Analysis and Perception lab, Aalborg University, Denmark\\%
$^2$Computer Vision Center, Universitat Autònoma de Barcelona, Spain\\%
$^3$Dept. of Mathematics and Informatics, Universitat de Barcelona, Spain\\%
{\tt\small gegeh@create.aau.dk}\qquad {\tt\small sescalera@ub.edu}\qquad {\tt\small tbm@create.aau.dk}}

\maketitle

\begin{abstract}

While there has been a growing research interest in developing out-of-distribution (OOD) detection methods, there has been comparably little discussion around how these methods should be evaluated. Given their relevance for safe(r) AI, it is important to examine whether the basis for comparing OOD detection methods is consistent with practical needs. In this work, we take a closer look at the go-to metrics for evaluating OOD detection, and question the approach of exclusively reducing OOD detection to a binary classification task with little consideration for the detection threshold. We illustrate the limitations of current metrics (AUROC \& its friends) and propose a new metric - Area Under the Threshold Curve (AUTC), which explicitly penalizes poor separation between ID and OOD samples. Scripts and data are available at \href{https://github.com/glhr/beyond-auroc}{\texttt{https://github.com/glhr/beyond-auroc}}
  
\end{abstract}

\section{Introduction}
\label{sec:intro}

When deployed out in the wild, computer vision systems may be faced with image content which they simply are not equipped to handle. For instance, a model trained to recognize certain types of skin lesions, once deployed in clinical practice, may encounter images with a different kind of skin condition, or images with no lesions at all~\cite{skin-lesions-ood-2022,derma-ood-2022}. Thus, it is not enough for models to make accurate predictions on the kind of content that they were trained on - they should also be able to express whether a new input is familiar enough to make a reliable prediction.

\begin{figure}[t]
    \centering
    \includegraphics[width=\linewidth]{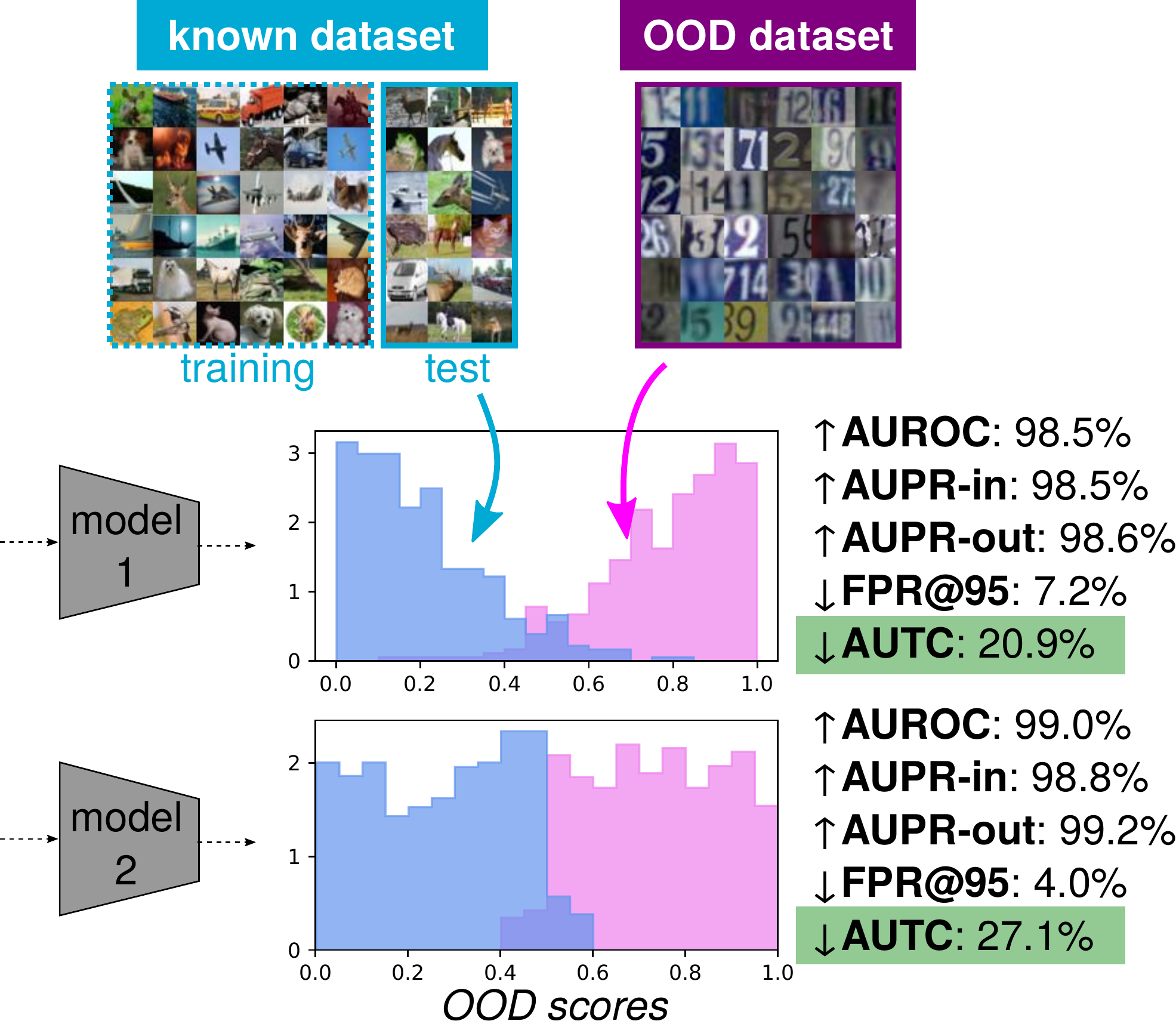}
    \caption{OOD evaluation setup illustrated with the CIFAR10~\cite{cifar-dataset-2009} vs.~SVHN~\cite{svhn-dataset-2011} pair. We visualize the OOD scores produced by 2 imaginary models as normalized histograms. Which model is better as an OOD detector? Popular metrics say model 2. We argue that more fine-grained metrics which take score distribution into account are needed for practical use, as they affect the choice of threshold for downstream decisions. We propose the AUTC metric which encourages separability between ID and OOD samples.}
    \label{fig:intro}
\end{figure}

The task of flagging images outside of a model's training domain is known as out-of-distribution (OOD) detection and is a growing line of research in computer vision~\cite{openood-benchmark-2022,unified-ood-survey-2022} with important implications for safe AI~\cite{ai-safety-problems-2016}. A broad range of methods have been proposed for equipping neural networks with OOD detection capabilities, ranging from uncertainty quantification~\cite{ue-deep-ensembles-2017,sngp-2020}, generative modelling~\cite{likelihood-ratios-ood-2019}, outlier exposure~\cite{ood-limits-2021}, to gradient-based~\cite{gradient-based-2021}, softmax-based~\cite{odin-2018}, distance-based~\cite{ood-knn-2022}, or energy-based~\cite{energy-based-ood-2020} approaches (among others). In this work, we abstract away from any specific OOD detection method, and rather focus on how OOD detection performance is quantitatively evaluated.

\cref{fig:intro} illustrates our problem setup. We consider generic models which were trained on an image recognition dataset and, given an input image, output an OOD score alongside their prediction - with a higher OOD score indicating that the image is more likely to be OOD. At evaluation time, the models are presented unseen in-distribution (ID) images (images within the training data distribution), along with OOD images from an unknown dataset. The scores are then aggregated and evaluated in terms of how well they can be used as a basis to distinguish between ID and OOD samples. This is the typical procedure in OOD benchmarks~\cite{openood-benchmark-2022,ood-benchmark-med-2022}. We then question the exclusive use of binary classification metrics (AUROC, AUPR, FPR \& co.) for quantitative comparison of different models, as they binarize OOD scores without considering their distribution or separability. Consider the example of \cref{fig:intro}, where \textit{model 2} outperforms \textit{model 1} across all standard metrics. Yet, looking at the distribution of scores, \textit{model 1} may be preferable in practice, as it achieves a much clearer separation between ID vs.~OOD samples, and allows more flexibility in the choice of threshold without drastic changes in detection performance. To expand on this intuition, we

\begin{itemize}[itemsep=2pt,topsep=2pt,parsep=1pt]
    \item briefly review the status quo in terms of metrics for evaluating OOD detection and identify confusing discrepancies in their definition across the literature (\cref{sec:existing-metrics})
    \item elaborate on some limitations of these metrics with the help of several illustrative examples (\cref{sec:problems}), and emphasize the need for a global detection threshold
    \item present an alternative view and new performance metric for evaluating OOD detectors with a focus on their downstream use, where separability between ID and OOD samples in terms of their OOD score is particularly important for the choice of a threshold (\cref{sec:alternative})
\end{itemize}

\paragraph{Related work} Within computer vision, recent surveys have outlined the most commonly used performance metrics in OOD detection and related tasks~\cite{generalized-ood-survey-2021,unified-ood-survey-2022} - this aligns with our brief review. However, to the best of our knowledge, there is little to no existing work discussing their limitations or considering possible extensions.

Complementary to this paper, \cite{ood-benchmarks-splitting-2021} discusses the design of OOD benchmarks for computer vision in terms of dataset splits, with the goal of minimizing semantic overlap between ID and OOD sets. ~\cite{ood-practical-eval-guidelines-2023} lays out practical guidelines and challenges for evaluating OOD detection when using medical data. In~\cite{derma-ood-2022}, the downstream implications of different kinds of mistakes are considered (e.g. flagging an OOD sample as ID vs.~ID sample as OOD) and modelled as a cost matrix in terms of model trustworthiness. Within the context of visual question answering, \cite{vqa-ood-2020} points out questionable but common practices in OOD benchmarks (e.g. tuning hyperparameters based on OOD performance), leading to misleadingly inflated performance.
\newpage
Beyond the realm of OOD detection,~\cite{failure-metareview-2021} reviews some of the pitfalls of benchmark-oriented machine learning, a major one being the use of simplified metrics which do not capture important differences between methods. In a similar spirit to our work,~\cite{metric-learning-reality-check-2020} questions the accuracy metrics reported in metric learning, as they fail to capture a notion of class separation.

\section{Performance metrics in OOD detection}\label{sec:existing-metrics}

\begin{figure}[h]
    \centering
    \includegraphics[width=0.9\linewidth]{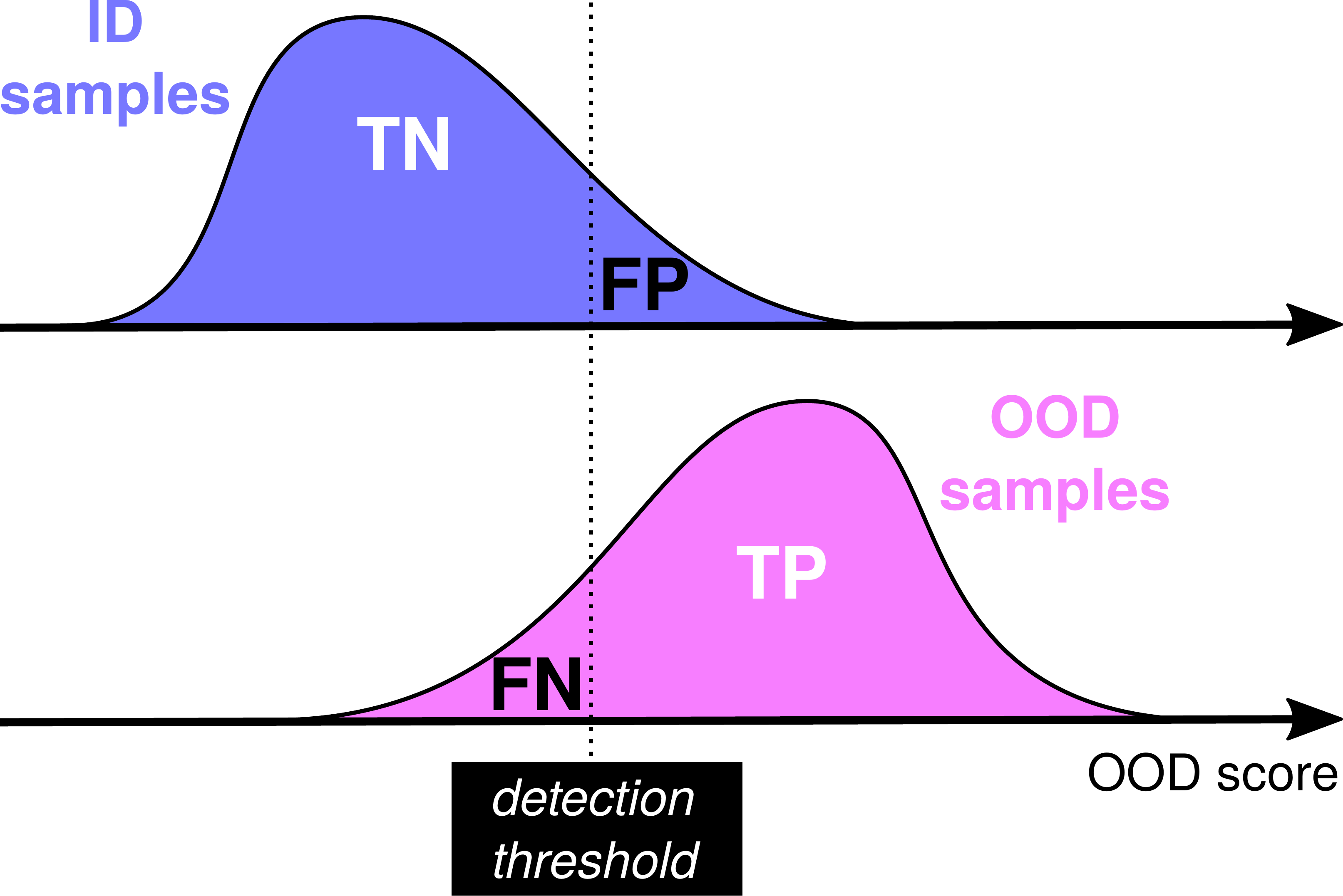}
    \caption{OOD scores are binarized based on a detection threshold.}
    \label{fig:binary-class-eval-histos}
\end{figure}

Overall, the consensus is to treat OOD detection as a binary classification task, where the predicted continuous OOD score is binarized and compared to a true label (positive if the test sample is OOD, negative otherwise - or vice-versa). The prediction is then either considered a True Positive (TP), True Negative (TN), False Positive (FP) or False Negative (FN) - as visualized in \cref{fig:binary-class-eval-histos}. Following the seminal work in~\cite{baseline-ood-2017}, the OOD detection literature has adopted the AUROC and AUPR as metrics of choice, thus bypassing the need to select a specific threshold. AUROC is often considered the main metric, and we did not find any works which did not report it. Alongside these, most works also report performance at a fixed detection threshold \cite{odin-2018,mahalanobis-ood-2018,residual-flow-2020,unsupervised-ood-2019,confidence-calib-class-ood-2018,openood-benchmark-2022,energy-based-ood-2020,likelihood-ratios-ood-2019,mos-ood-2021,ss-learning-ood-2020,scaling-ood-realworld-2022,derma-ood-2022,ood-limits-2021}. We briefly present each metric below.

\paragraph{Fixed-threshold metrics} These consider performance at a specific operating point. The FPR@TPR metric measures the false positive rate for a given true positive rate - typically chosen to be 95\%~\cite{energy-based-ood-2020,openood-benchmark-2022,unsupervised-ood-2019,mos-ood-2021}, or sometimes 80\%~\cite{likelihood-ratios-ood-2019}. In a similar vein, some works instead report the TNR@95 (true negative rate at 95\% TPR)~\cite{mahalanobis-ood-2018,confidence-calib-class-ood-2018}. For an ideal detector, the TNR is 100\% while the FPR is 0\%.

\begin{figure*}[b]
    \centering
    \includegraphics[width=\textwidth]{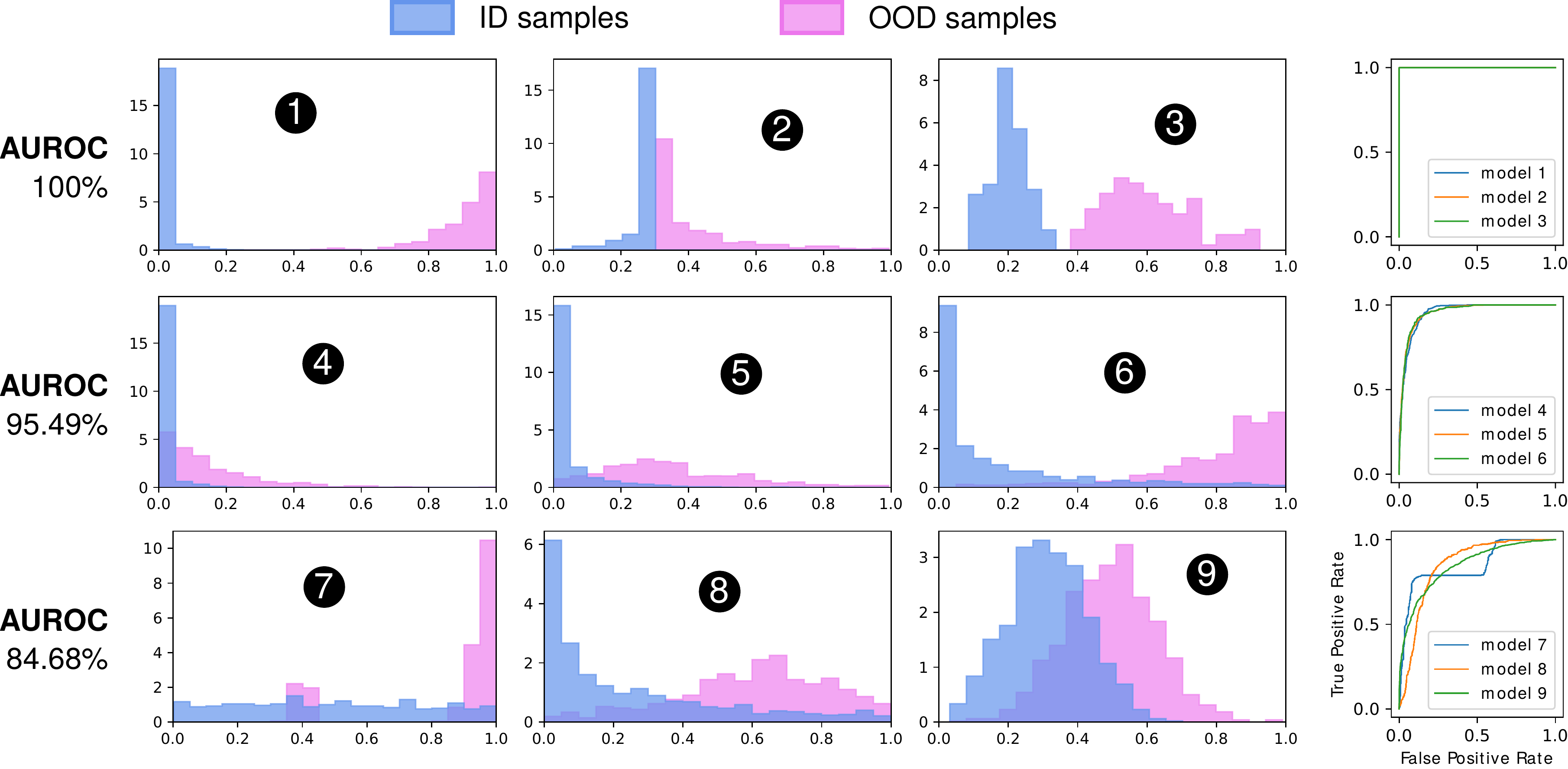}
    \caption{Histograms (normalized) of OOD scores for ID vs.~OOD samples produced by 9 imaginary models. The models on a given row achieve the same (rounded to 4 decimal places) OOD performance in terms of AUROC. On the right are ROC curves.}
    \label{fig:auroc-rows}
\end{figure*}

\paragraph{Threshold-independent metrics} These summarize OOD detection performance with a sliding threshold. The AUROC measures the area under the Receiver Operating Characteristic curve, obtained by plotting the TPR as a function of the FPR. It can be interpreted as the probability that a positive sample is assigned a higher score than a negative sample - an AUROC of 100\% indicates perfect separation, while an AUROC of 50\% indicates full overlap (uninformative/random detector). The AUPR measures the area under the curve obtained by plotting recall (R) as a function of precision (P). Unlike AUROC, it is sensitive to sample size and the choice of positive class, hence it is common to distinguish between AUPR-in (ID samples considered positive) and AUPR-out (OOD samples considered positive)~\cite{odin-2018}. 

\paragraph{Detection error/accuracy} Some works~\cite{confidence-calib-class-ood-2018} report this as the best detection accuracy/error across all possible thresholds, while others report it for a fixed TP rate (e.g. 95\% in ~\cite{odin-2018}). As discussed in~\cite{baseline-ood-2017}, a drawback of using detection accuracy as a metric is that it is skewed by class imbalance (ie. it assumes an equal number of ID and OOD samples).

\paragraph{Positive or negative, that is the question} We note a mismatch in the literature in terms of which class (ID or OOD) is considered positive or negative for the computation of metrics. While one set of papers treat ID samples as the positive class~\cite{baseline-ood-2017,odin-2018,confidence-calib-class-ood-2018,openood-benchmark-2022}, another instead treats OOD samples as positive~\cite{ss-learning-ood-2020,mos-ood-2021,vim-ood-2022}. Some works fail to mention this definition altogether~\cite{sngp-2020}. While this may only seem like a minor difference of terminology, it affects the AUPR and FPR computation - both widely used metrics. For instance, the FPR@95 metric reported in the recent benchmark~\cite{openood-benchmark-2022} vs.~the one reported in~\cite{vim-ood-2022} are in fact different metrics due to this mismatch in class definition. Special care should therefore be taken to avoid such inconsistencies. In the context of this paper, we consider OOD samples to be positive unless otherwise specified.

\pgfplotsset{cycle list/Paired-9}
\begin{figure*}[t]
    \centering
    \begin{tikzpicture}
    \begin{axis}[
            ybar,
            width=\textwidth,
            height=4.5cm,
            bar width=.25cm,
            ybar=1pt,
            legend style={at={(0.5,1.25)},
                anchor=north,legend columns=-1,
                draw=none,
                /tikz/every even column/.append style={column sep=0.2cm}},
           xticklabels from table={\scores}{metric},
           xtick=data,
            tick pos=left,
            nodes near coords,
            nodes near coords align={horizontal},
            nodes near coords style={rotate=90, font=\small},
            ymin=0,ymax=119,
            ylabel={Score \%},
                enlarge y limits = {value = .25, upper},
                enlarge x limits = 0.15,
            every axis plot/.append style={fill},
            cycle multi list={Paired-9},
        ]
        
        \addplot table[x expr=\coordindex,y=model1]{\scores};
        \addplot table[x expr=\coordindex,y=model2]{\scores};
        \addplot table[x expr=\coordindex,y=model3]{\scores};
        \addplot table[x expr=\coordindex,y=model4]{\scores};
         \addplot table[x expr=\coordindex,y=model5]{\scores};
          \addplot table[x expr=\coordindex,y=model6]{\scores};
          \addplot table[x expr=\coordindex,y=model7]{\scores};
         \addplot table[x expr=\coordindex,y=model8]{\scores};
          \addplot table[x expr=\coordindex,y=model9]{\scores};
        \legend{1,2,3,4,5,6,7,8,9}
    \end{axis}
    
\end{tikzpicture}

\caption{Performance of the 9 imaginary models in terms of standard metrics. Models are numbered according to \cref{fig:auroc-rows}.}
\label{fig:standardmetrics-bar}
\end{figure*}
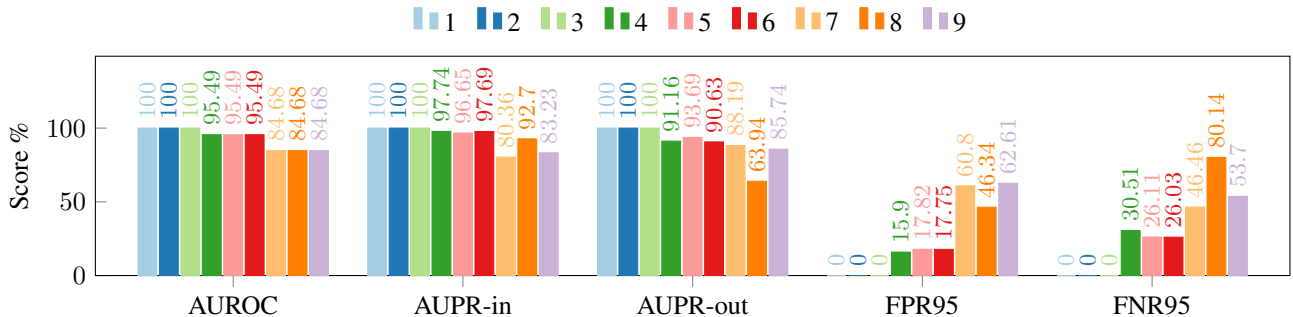

\section{What's the problem?}\label{sec:problems}

\subsection{Let histograms (s)peak for themselves}

Assessing the performance of OOD detection methods purely in terms of binary classification metrics reduces the evaluation to a comparison of ratios between the number of TP, TN, FP and FN predictions, without considering how OOD scores are distributed. That is, two models with the same performance may differ widely in terms of how clearly they separate ID from OOD samples. In a similar spirit to \cref{fig:intro}, we illustrate this effect in \cref{fig:auroc-rows}, where we simulate OOD scores produced by 9 imaginary models (which were trained and evaluated on the same imaginary data), such that models on a given row exhibit near-identical performance in terms of AUROC. Other standard metrics for each model are reported in \cref{fig:standardmetrics-bar}. 

\paragraph{Is high performance all we need?} The first row in \cref{fig:auroc-rows} shows examples of ``perfect'' OOD detection models in terms of standard metrics, with no overlap between the two classes (such high detection performance is not unheard of - for instance, the method in~\cite{dhms-ood-2022} achieves 100\% AUROC and AUPR on the CIFAR10/100 vs.~SVHN pair). Despite their identical performance, if tasked with picking a model to deploy in a practical application, one would most likely prefer model 1 over model 2 due to the clear separation between ID and OOD samples. Model 2 inspires less confidence outside of a benchmark scenario (where we only consider images present in the test set), as its performance is extremely sensitive to the choice of threshold. Yet, none of the standard metrics capture this distinction.

In the second row, model 4 is a prime example of a model achieving what is considered an ``excellent'' OOD detection performance by common standards~\cite{baseline-ood-2017}, but very poor separation between scores assigned to ID vs.~OOD samples. Trying to select a suitable operating point for this model would be less straightforward than for model 6 - if set slightly too low, many inputs would be wrongly flagged as OOD, but if slightly higher, a large portion of OOD samples would be missed. Across the performance metrics in~\cref{fig:standardmetrics-bar}, FNR@95 is the only metric penalizing model 4.

Lastly, the models in the third row would be considered ``good'' OOD detectors based on their AUROC of almost 85\%~\cite{baseline-ood-2017}. Model 7 exhibits quite undesirable behaviour, with OOD scores for ID samples uniformly distributed across the whole range, yet it achieves the best performance in terms of AUPR-out and FNR@95. Models 8 and 9 have a clearer separation between ID vs.~OOD samples, with a sensible threshold lying around 0.4. The large differences between AUPR-in vs.~AUPR-out and FPR@95 vs.~FNR@95 results for model 8 in~\cref{fig:standardmetrics-bar} also highlights that reporting only one ``side'' of these metrics can give a misleading picture of performance.

\vspace{-1em}

\paragraph{In a nutshell} As we have shown with the examples of \cref{fig:auroc-rows}, standard metrics are sensitive to the amount of \textit{overlap} between scores assigned to ID vs.~OOD samples, but are blind to the level of \textit{separation} between them. Indeed, a model achieving perfect performance in terms of AUROC or AUPR only means that there exists at least \textit{one} threshold for which the FPR and FNR are 0.  We argue that having a wide range of sensible thresholds to choose from is a desirable property for an OOD detector.

\begin{figure}[t]
    \centering
    \includegraphics[width=0.7\linewidth]{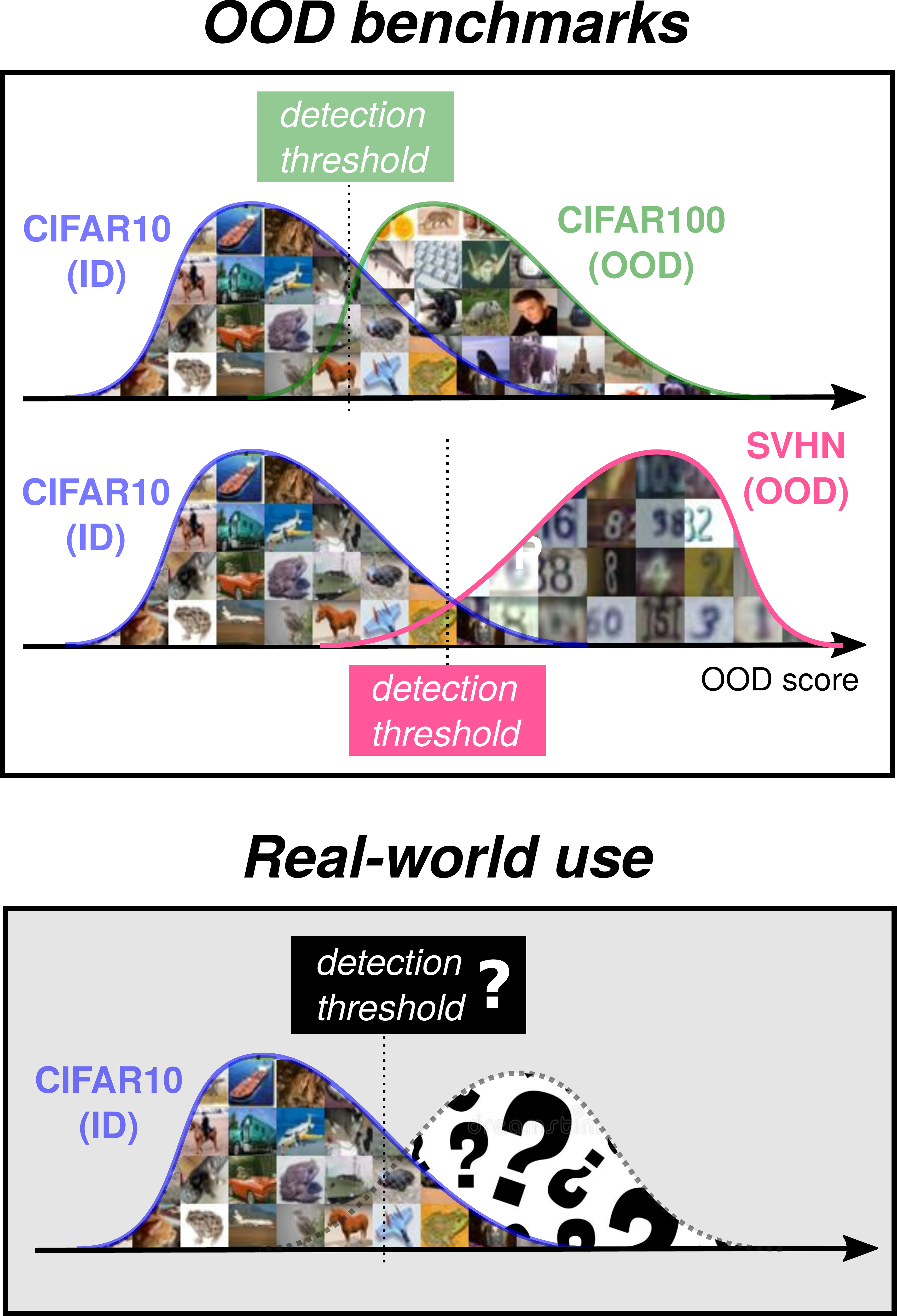}
    \caption{In a benchmark setting, once the model has been trained (on CIFAR10 in this example), the threshold for metrics is often set based on performance on each individual OOD dataset (e.g. CIFAR100 or SVHN). However, deployment of the model in a practical setting requires the choice of a single sensible threshold which is not tailored to a specific OOD dataset.}
    \label{fig:benchmark-vs-real}
\end{figure}

\subsection{About that threshold}

\begin{figure*}[b]
    \centering
    \includegraphics[width=\textwidth]{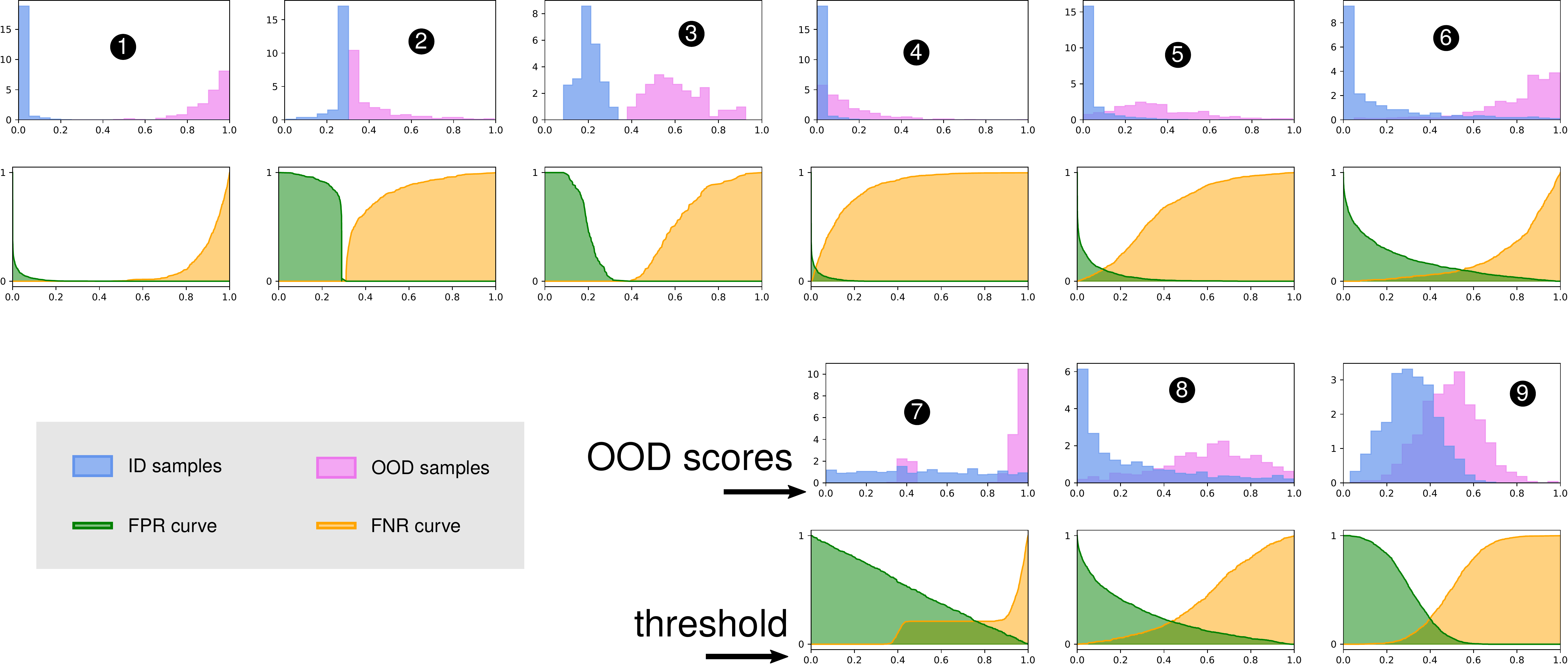}
    \caption{Visualization of the FPR and FNR vs.~threshold curves used to compute our proposed metric. The corresponding histogram of each model is shown above them for reference.}
    \label{fig:autc-rows}
\end{figure*}

Besides the fact that the metrics themselves only tell part of the story, another practical concern is that current evaluation of OOD detectors essentially considers each ID vs.~OOD dataset pair as its own classification task. For example, a model trained on CIFAR10, will then independently be evaluated on CIFAR10 vs.~SVHN and CIFAR10 vs.~CIFAR100 (and often several other OOD datasets~\cite{ss-learning-ood-2020}), with the above-mentioned metrics reported for each pair. Thus, even when employing a fixed-threshold metric, the threshold may vary across the different OOD datasets in the evaluation - as illustrated in~\cref{fig:benchmark-vs-real}. We argue that this approach is fundamentally misaligned with a real-use setting, for which a single detection threshold has to be chosen.



\section{What next?}\label{sec:alternative}

We suggest a shift of perspective: consider the OOD scores not just as the output of a classifier, but as an \textit{input} to a decision function which has to determine whether to discard the model's prediction on the original task. The selection of a threshold for this decision function will determine the actual OOD detection performance for new images, and is therefore a safety-critical design choice. 

With the threshold in mind, we present a new performance metric along with some recommendations for the evaluation of OOD detectors.

\subsection{Enter a new metric}

If model performance at evaluation time (that is, for a single ID vs.~OOD dataset pair) is extremely sensitive to the choice of threshold, then it is difficult to imagine robust performance at run-time, where the diversity of OOD samples is expected to increase. Yet, standard area-based performance metrics do not capture the relation between a change of threshold and a change in performance.

To address this gap and quantify important differences between models which are not caught by existing metrics, we therefore propose a new metric, the Area Under the Threshold Curve (AUTC), which explicitly penalizes poor separation between ID and OOD samples. Increased separability is sometimes mentioned as a desired property for OOD detectors in the state of the art~\cite{odin-2018}, but has not been previously quantified.

In contrast with the ROC curve or PR curve, our metric is based on a visualization which explicitly shows the effect of the threshold on OOD detection performance by plotting the FPR and FNR of the detector as a function of the detection threshold. As shown in  \cref{fig:autc-rows}, not only are these plots convenient for visually choosing a threshold, they also reveal stark differences between models - even those with perfect AUROC - in terms of how quickly the performance degrades when moving away from the curves' crossing point. They also implicitly combine both a measure of performance and separability: the lower the crossing point, the higher the OOD detection performance with a good choice of threshold, and the closer the curves are glued to the Y axis on each side, the more ID and OOD scores are concentrated at opposite sides of the score ranges.

Notably, as the area under the FPR curve (AUFPR) and under the FNR curve (AUFNR) shrink, we approach an ideal detector which assigns a score of 0 to all ID samples, and 1 to all OOD samples. 

\paragraph{AUTC metric} We summarize these curves as a single metric by combining the AUFPR and AUFNR, and averaging them to obtain a single value within $[0,1]$:

\[
\textrm{AUTC} = \frac{\textrm{AUFPR} + \textrm{AUFNR}}{2}
\]

\begin{figure*}[t]
        \centering
    \begin{tikzpicture}
    \begin{axis}[
            ybar,
            width=0.7\linewidth,
            height=4.5cm,
            bar width=.24cm,
            ybar=1pt,
            legend style={at={(1.2,0.7)},
                anchor=north,legend columns=3,
                draw=none,
                /tikz/every even column/.append style={column sep=0.2cm}},
           xticklabels from table={\ourscores}{metric},
           xtick=data,
            tick pos=left,
            nodes near coords,
            nodes near coords align={horizontal},
            nodes near coords style={rotate=90, font=\small},
            ymin=0,ymax=100,
            ylabel={Score \%},
             enlarge y limits = {value = .25, upper},
                enlarge x limits = 0.3,
            every axis plot/.append style={fill},
            cycle multi list={Paired-9},
            y filter/.code={\pgfmathparse{#1*100}\pgfmathresult}
        ]
        
        \addplot table[x expr=\coordindex,y=model1]{\ourscores};
        \addplot table[x expr=\coordindex,y=model2]{\ourscores};
        \addplot table[x expr=\coordindex,y=model3]{\ourscores};
        \addplot table[x expr=\coordindex,y=model4]{\ourscores};
         \addplot table[x expr=\coordindex,y=model5]{\ourscores};
          \addplot table[x expr=\coordindex,y=model6]{\ourscores};
          \addplot table[x expr=\coordindex,y=model7]{\ourscores};
         \addplot table[x expr=\coordindex,y=model8]{\ourscores};
          \addplot table[x expr=\coordindex,y=model9]{\ourscores};
           \legend{1,2,3,4,5,6,7,8,9}
        
    \end{axis}
    
\end{tikzpicture}

    \caption{Our proposed OOD performance metrics for the same 9 models. Lower performance is better.}
    \label{fig:ourmetrics-bar}
\end{figure*}
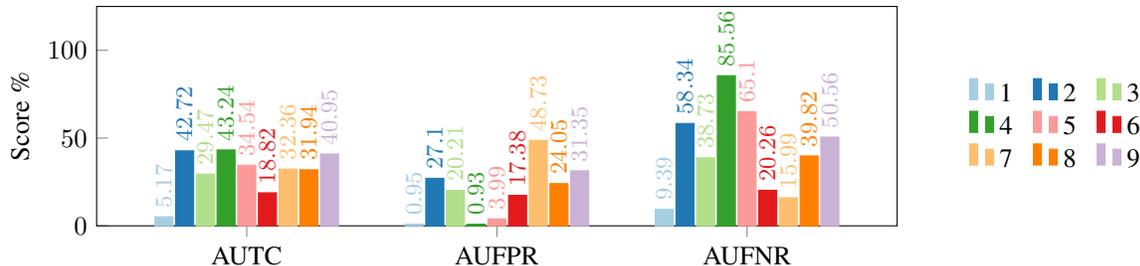

In practice, the areas can be computed via the trapezoidal rule, as is commonly done for the AUROC or AUPR.

Some properties of the AUTC:
\begin{itemize}[itemsep=1pt,topsep=2pt,parsep=1pt]
    \item a lower value is better: it is equal to 1 for the worst detector with complete separation, 0.5 for a random detector (no separation), and 0 for a perfect detector with complete separation.
    \item it is not sensitive to sample size or to the definition of positive vs.~negative class.
    \item it assumes OOD scores between 0 and 1. 
\end{itemize}

In \cref{fig:ourmetrics-bar}, we show the AUTC computed for the 9 models, as well as the corresponding AUFPR and AUFNR. This comparison reveals significant differences compared to the standard metrics from ~\cref{fig:standardmetrics-bar}. This time, in terms of AUTC, Models 1 and 6 clearly stand out compared to the other models, as the metric encourages strong separability, while model 4 obtains the worst score despite its AUROC of over 95\%. Looking at the AUFPR and AUFNR allows us to separately quantify the spread and distance from 0/1 for ID vs.~OOD samples. For instance, Models 4-6 have a significantly lower AUFPR than Models 7-8, as the scores assigned to ID samples are more concentrated around 0.

\subsection{Other considerations for evaluation}

We note several limitations of the AUTC metric:
\begin{enumerate}[itemsep=2pt,topsep=2pt,parsep=1pt]
    \item The AUTC encourages strong separability - that is, it encourages OOD scores to be concentrated around 0 for all ID samples, and 1 for all OOD samples. This behaviour may not be desirable if one instead wishes the OOD scores to be well-calibrated (that is, for the OOD score to be a good indicator of the \textit{probability} of a sample being OOD).
    \item Since it simply averages the AUFPR and AUFNR, the AUTC gives equal weight to false negatives and false positives. In practice, the cost of these two types of mistakes may not be symmetrical (as discussed in~\cite{derma-ood-2022}). We therefore recommend separately reporting the AUFPR and AUFNR, and/or giving them different weights in the AUTC computation based on severity.
    \item As is the case for other curves parametrized by a threshold~\cite{reducing-network-agnostophobia-2018}, the AUTC is sensitive to transformations of the OOD scores. 
\end{enumerate}

Furthermore, much like AUROC and AUPR, our proposed metric is a summary metric covering all possible thresholds. When comparing OOD detection methods, it should be accompanied by a fixed-threshold metric (e.g. by reporting FPR and FNR at a specific operating point). However, contrary to common practice, we emphasize that this fixed threshold should not be tailored to the OOD datasets used in the final evaluation. It should instead be tuned either on the ID dataset, or on a separate validation OOD set which is not used during final evaluation - as we cannot assume to know the distribution of unseen OOD samples. During evaluation, to reflect real-world conditions, the same threshold should be used across all OOD datasets when reporting fixed-threshold metrics.



\section{A concrete example}

Moving away from imaginary models and synthetically-generated data, we demonstrate our approach on real OOD models and ID vs.~OOD dataset pairs. We select 2 models from the state of the art trained on the good old CIFAR10~\cite{cifar-dataset-2009} dataset, and compare their OOD detection performance on multiple unseen datasets (CIFAR100, tinyImageNet~\cite{tiny-imagenet-2015}, SVHN~\cite{svhn-dataset-2011}, and LSUN~\cite{lsun-dataset-2015}). We briefly present the models below, and refer to the original papers for details:
\begin{enumerate}[itemsep=0pt]
    \item Out-of-DIstribution detector for Neural networks (ODIN) from~\cite{odin-2018} applies temperature scaling and input perturbations to a pre-trained neural network. The OOD score is based on the maximum softmax probability. We use the DenseNet model weights from the official code repository\footnote{\url{https://github.com/facebookresearch/odin}}.
    \item Spectral-normalized Neural Gaussian Process (SNGP) from ~\cite{sngp-journal-22} combines a distance-preserving feature extractor with an approximate Gaussian process as output layer. The OOD score is taken as the Dempster-Shafer metric. We train a model following a third-party PyTorch implementation\footnote{\url{https://github.com/y0ast/DUE}}.
\end{enumerate}

\begin{table*}[t]
\centering
\begin{subtable}[h]{0.48\textwidth}
\centering
\begin{tabular}{@{}lllcccc@{}}
\toprule
  & & & \multicolumn{3}{c}{\textbf{OOD datasets}} \\
  & & & \begin{tabular}[c]{@{}c@{}}tiny\\ Imagenet\end{tabular} & \multicolumn{1}{c}{LSUN} & \multicolumn{1}{c}{SVHN} \\ \midrule
& \multicolumn{2}{r}{AUROC $\uparrow$} & 99.11 & 97.89 & 89.90 \\
\rule{0pt}{0.5cm}

 & \multicolumn{2}{r}{AUFPR $\downarrow$} & \multicolumn{3}{c}{------------  45.17 ------------}  \\
 & \multicolumn{2}{r}{AUFNR $\downarrow$} & 21.97 & 25.24 & 35.11  \\
 & \multicolumn{2}{r}{\textbf{AUTC} $\downarrow$} & \textbf{33.57} & \textbf{35.21} & \textbf{40.41} \\

\rule{0pt}{0.5cm}

& \multirow{2}{*}{\footnotesize{@test}} & FPR $\downarrow$ & \multirow{2}{*}{4.71} & \multirow{2}{*}{4.40} & \multirow{2}{*}{18.51} \\
& & FNR $\downarrow$ &  &  &  \\

\rule{0pt}{0.5cm}

 & \multirow{2}{*}{\footnotesize{@95TNR}} & FPR $\downarrow$ & \multicolumn{3}{c}{------------ 5.00 ------------} \\
 &   & FNR $\downarrow$ & 4.30 & 11.40 & 50.92 \\

 \rule{0pt}{0.5cm}
 
 & \multirow{2}{*}{\footnotesize{@val}} & FPR $\downarrow$ & \multicolumn{3}{c}{------------ 19.56 ------------} \\
 &   & FNR $\downarrow$ & 0.46 & 1.88 & 17.00 \\

 \bottomrule
\end{tabular}
\caption{ODIN}
 \end{subtable}\hfill
\begin{subtable}[h]{0.48\textwidth}
\centering
\begin{tabular}{@{}lllcccc@{}}
\toprule
 &  & & \multicolumn{3}{c}{\textbf{OOD datasets}} \\
 &  & & \begin{tabular}[c]{@{}c@{}}tiny\\ Imagenet\end{tabular} & \multicolumn{1}{c}{LSUN} & \multicolumn{1}{c}{SVHN} \\ \midrule
 & \multicolumn{2}{r}{AUROC $\uparrow$} & 97.38 & 99.09 & 99.06 \\
\rule{0pt}{0.5cm}

   & \multicolumn{2}{r}{AUFPR $\downarrow$} & \multicolumn{3}{c}{------------ 2.85 ------------}  \\
 & \multicolumn{2}{r}{AUFNR $\downarrow$} & 49.98 & 29.41 & 33.79 \\
 & \multicolumn{2}{r}{\textbf{AUTC} $\downarrow$} & \textbf{26.42} & \textbf{16.13} & \textbf{18.32} \\

\rule{0pt}{0.5cm}

& \multirow{2}{*}{\footnotesize{@test}} & FPR $\downarrow$ & \multirow{2}{*}{8.36} & \multirow{2}{*}{5.01} & \multirow{2}{*}{4.97} \\
&  & FNR $\downarrow$ &  &  &  \\

\rule{0pt}{0.5cm}
 
  & \multirow{2}{*}{\footnotesize{@95TNR}} & FPR $\downarrow$ & \multicolumn{3}{c}{ ------------ 5.00 ------------} \\
 &   & FNR $\downarrow$ & 14.11 & 5.02 & 4.95 \\

 \rule{0pt}{0.5cm}
 
 & \multirow{3}{*}{\footnotesize{@val}} & FPR $\downarrow$ & \multicolumn{3}{c}{------------ 16.22 ------------} \\
 &   & FNR $\downarrow$ & 3.19 & 0.89 & 0.77 \\
\bottomrule
\end{tabular}
\caption{SNGP}
\end{subtable}
\caption{Quantitative comparison of the 2 models' OOD detection performance. Scores are in percentages, and the proposed AUTC metric is highlighted in bold. Note that when fixing a global threshold, the FPR is the same across all OOD datasets as it only depends on the distribution of OOD scores for the ID data (CIFAR10).}
\label{tab:due-odin}
\end{table*}

Note that the purpose of this experiment is not to pit one method against another, but rather to show our proposed metric and evaluation procedure in action.

\begin{figure}[t]
\centering
\begin{subfigure}{0.45\linewidth}
\centering
    \includegraphics[width=\textwidth]{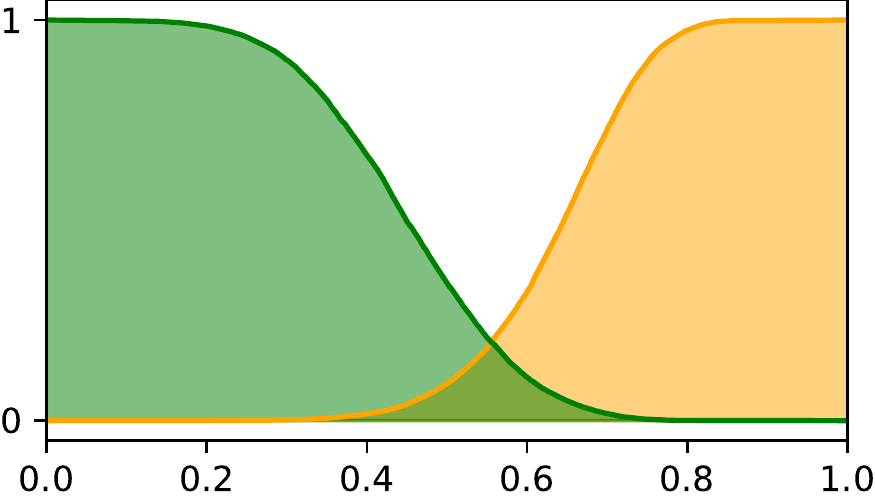}
    
    \vspace{0.3cm}
    
    \includegraphics[width=\textwidth]{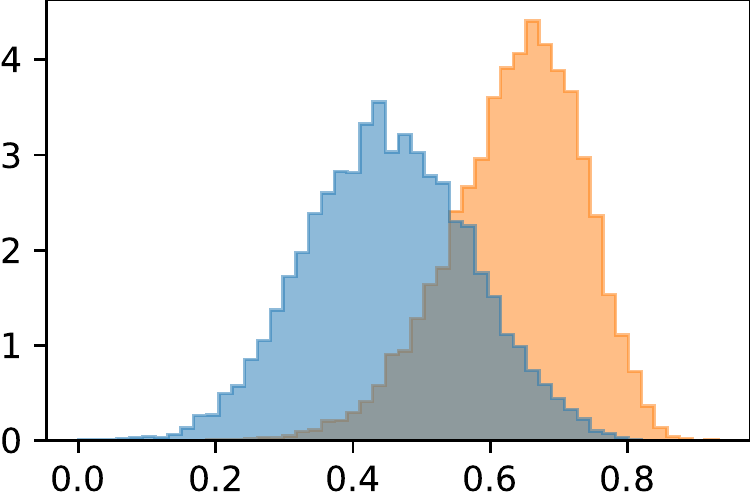}
     \caption{ODIN}
     \end{subfigure}\hfill
    \begin{subfigure}{0.45\linewidth}
    \centering
    \includegraphics[width=\textwidth]{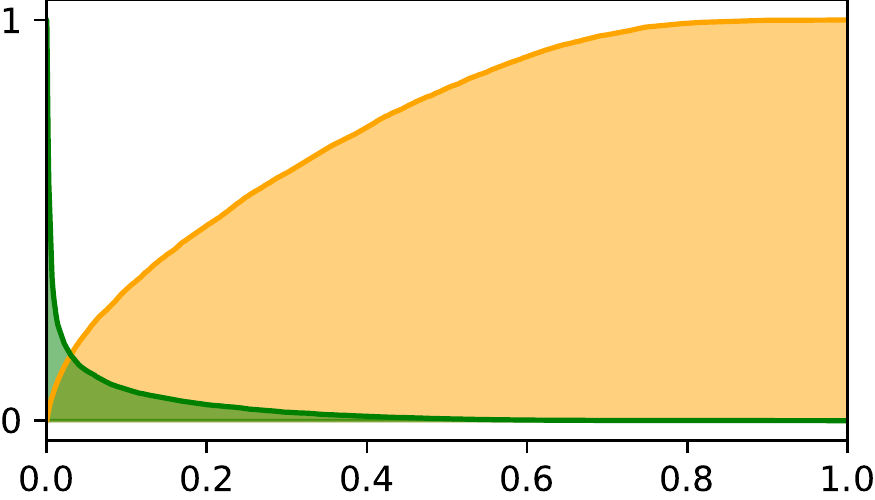}
    
    \vspace{0.3cm}
    
    \includegraphics[width=\textwidth]{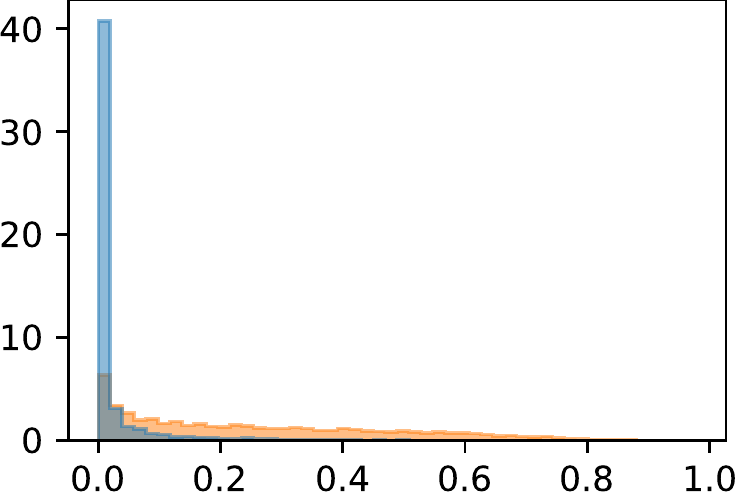}
    \caption{SNGP}
    \end{subfigure}

\includegraphics[width=0.4\linewidth]{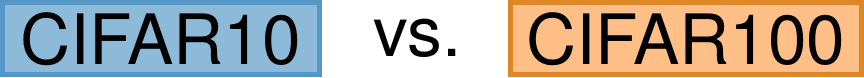}

    \caption{FPR (green) \& FNR (yellow) curves and normalized histograms of OOD scores predicted by the 2 models on CIFAR10 (ID dataset) vs.~CIFAR100 (val OOD dataset).}
    \label{fig:histos-sngp-odin}
\end{figure}

\paragraph{More plots} We use CIFAR100 as a validation dataset for the detection threshold (as it is ``closer'' to the training set in terms of appearance than the others), and the rest of the OOD datasets for the final evaluation. As visualized in \cref{fig:histos-sngp-odin}, altough they exhibit comparable detection performance (similar amount of overlap between the FPR and FNR curves), the models differ widely in terms of how their OOD scores are distributed for ID vs.~OOD samples. For ODIN, a reasonable threshold lies around 0.55, while SNGP concentrates its scores for ID samples around 0. As the threshold increases from the crossing point, the FNR increases more rapidly for ODIN than SNGP.

\paragraph{Finally a table} We summarize the quantitative results in~\cref{tab:due-odin}. For each model and ID vs.~OOD pair, we report the AUROC (main standard metric), AUTC (our proposed metric) as threshold-independent performance metrics. We also measure the FNR (probably of misclassifying an OOD sample as ID) for several fixed thresholds:
\begin{itemize}[itemsep=0pt,topsep=1pt,parsep=1pt]
    \item \textbf{@test} - the point at which the FPR and FNR on the OOD set are equal (this threshold is specific to each OOD dataset). We include this as reference, to show ideal performance.
    \item \textbf{@95TNR} - the point at which the TNR is at least 95\%. This threshold only depends on the distribution of ID scores (thus stays the same across OOD datasets).
    \item \textbf{@val} - the point at which the FPR and FNR on the validation dataset CIFAR100 are equal (also stays the same across OOD datasets).
\end{itemize}

Looking at the threshold-independent metrics (AUROC to AUTC in~\cref{tab:due-odin}), both models achieve similar and high (potential for) OOD detection performance, except for SVHN where there is a difference of 10 percentage points in AUROC between ODIN and SNGP. Our AUTC metric correlates with AUROC performance for a given model, while also indicating that SNGP produces better separability between ID vs.~OOD samples. Note that the AUFPR is constant across OOD datasets as it only depends on the distribution of ID samples - SNGP has a much lower AUFPR due to a strong concentration of OOD scores around 0 for ID samples.

\begin{figure}[h]
\centering
\begin{subfigure}{0.45\linewidth}
\centering
    \includegraphics[width=\textwidth]{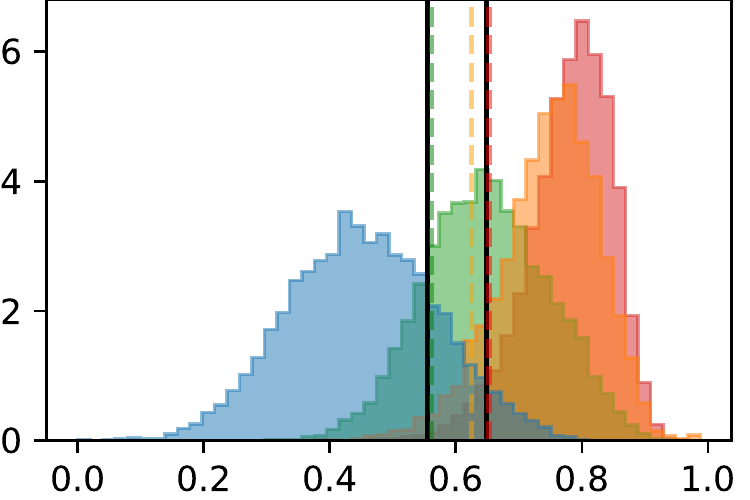}
     \caption{ODIN}
     \end{subfigure}\hfill
    \begin{subfigure}{0.45\linewidth}
    \centering
    \includegraphics[width=\textwidth]{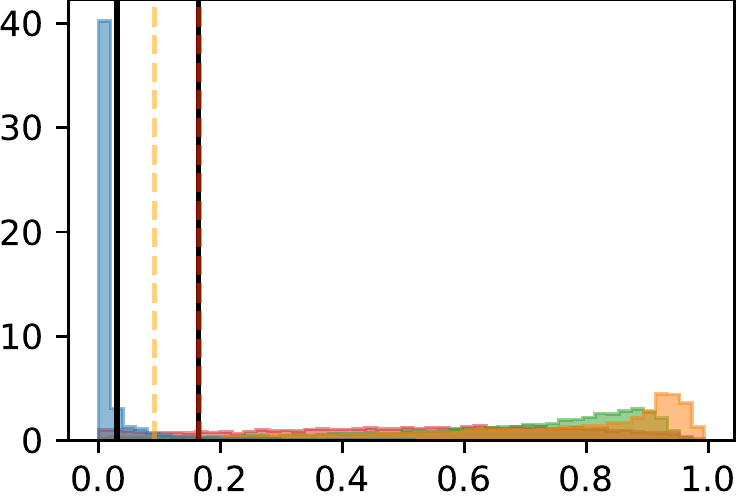}
    \caption{SNGP}
    \end{subfigure}

\includegraphics[width=0.6\linewidth]{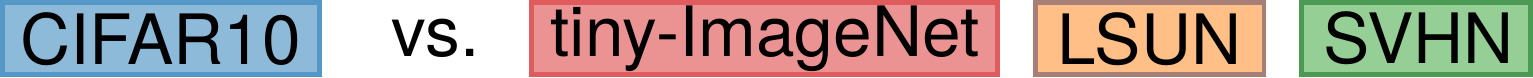}

    \caption{Normalized histograms of OOD scores for the test OOD pairs. The global (black) and @test (dashed, colored by dataset) thresholds are shown as vertical lines.}
    \label{fig:histos-thresh}
\end{figure}

The threshold-specific results (FPR and FNR in~\cref{tab:due-odin}) show how significantly the choice of threshold can impact performance. The performance at @test is unrealistic in practice, as it assumes that the threshold can be adjusted for each OOD dataset - as shown in \cref{fig:histos-thresh}. Fixing a global threshold based on the ID dataset scores (@95TNR) or the validation OOD set (@val) widens the performance gap across different ID vs.~OOD pairs. When shifting the global threshold from one to the other, we see larger fluctuations in performance on OOD datasets with a higher AUTC (SVHN for ODIN, and tinyImagenet for SNGP). 

\section{Zooming out}

\paragraph{Beyond CIFAR} Throughout the illustrative examples of this paper, we have considered the context of image classification, as this is the most common setting in OOD benchmarks for computer vision~\cite{openood-benchmark-2022}. However, our analysis of OOD metrics extends well beyond this setting, as it abstracts away from any particular input modality (images, audio, point clouds...) or main task (classification, regression..), as well as how or at what level of granularity (e.g. image-level, pixel-level...) the OOD scores are produced.

\paragraph{Future work} We have emphasized the need for evaluating OOD detection with the choice of a global threshold in mind, as this choice would have to made for any practical application: samples with an OOD score above this threshold are flagged as OOD, allowing the system to fallback to a safe strategy (e.g. requesting human input) rather than allowing the model to make a prediction. Investigating how a ``good'' global threshold should be chosen (without assuming the distributions of OOD datasets are known) and whether ID vs.~OOD separability indeed translates to more robust OOD detection performance are important directions for future research, as well as developing methods which incorporate the choice of threshold in the model design itself.

\section{Conclusion}

In this work, we have focused on the quantitative evaluation of OOD detectors, highlighting that current performance metrics can lead to misleading comparisons between methods due to a terminology mismatch in the OOD detection literature, and can obfuscate some important differences between models such as their ability to produce clearly separated scores for ID vs.~OOD samples. With concrete examples, we have shown that achieving a high performance in terms of AUROC is only the first step towards utilizing OOD detection in practical settings, and that the choice of a detection threshold should be treated as an important hyperparameter rather than an afterthought. We have presented a new metric which can serve as a complementary basis for comparing OOD detection models in terms of how well they separate ID from OOD samples by OOD score. We hope that this paper serves as a starting point to encourage further discussion around how OOD detection methods should be evaluated to align with the goals of practical and safe AI.

\section{Acknowledgements}

This work was supported by the Danish Data Science Academy, which is funded by the Novo Nordisk Foundation (NNF21SA0069429) and VILLUM FONDEN (40516). This research was also supported by the Pioneer Centre for AI, DNRF grant number P1. Last but not least, special thanks to cats.

{\small
\bibliographystyle{ieee_fullname}
\bibliography{refs}
}

\end{document}